\documentclass[runningheads]{llncs}
\usepackage[T1]{fontenc}
\usepackage{booktabs}
\usepackage{multirow}
\usepackage{tabularx}
\usepackage[linesnumbered,ruled,vlined]{algorithm2e}
\usepackage[table]{xcolor}
\usepackage{booktabs}
\usepackage{amsmath,amssymb}
\usepackage{pifont} 
\usepackage{graphicx}
\usepackage{colortbl}
\usepackage{lipsum} 
\usepackage{tgheros}
\usepackage{graphicx,verbatim}

\newcolumntype{Y}{>{\centering\arraybackslash}X}
\newcommand{\cmark}{\textcolor[HTML]{0B6B91}{\ding{51}}} 
\newcommand{\xmark}{\textcolor[HTML]{282B34}{\ding{55}}} 
\newcommand{\textsb}[1]{{\fontseries{sb}\selectfont #1}}

\begin{document}
\title{Revisit the Stability of Vanilla Federated Learning Under Diverse Conditions}
\author{Youngjoon Lee\inst{1}\and
Jinu Gong\inst{2}\and
Sun Choi\inst{3}\and
Joonhyuk Kang\inst{1}}

\authorrunning{Y. Lee et al.}
%
\institute{School of Electrical Engineering, KAIST, South Korea\and
Department of Applied AI, Hansung University, South Korea \and
Data Center and AI Group, Intel Corporation, United States\\
\email{yjlee22@kaist.ac.kr}, \email{jkang@kaist.ac.kr}
}

\maketitle
\begin{abstract}
Federated Learning (FL) is a distributed machine learning paradigm enabling collaborative model training across decentralized clients while preserving data privacy.
In this paper, we revisit the stability of the vanilla FedAvg algorithm under diverse conditions. 
Despite its conceptual simplicity, FedAvg exhibits remarkably stable performance compared to more advanced FL techniques. 
Our experiments assess the performance of various FL methods on blood cell and skin lesion classification tasks using Vision Transformer (ViT).
Additionally, we evaluate the impact of different representative classification models and analyze sensitivity to hyperparameter variations.
The results consistently demonstrate that, regardless of dataset, classification model employed, or hyperparameter settings, FedAvg maintains robust performance. 
Given its stability, robust performance without the need for extensive hyperparameter tuning, FedAvg is a safe and efficient choice for FL deployments in resource-constrained hospitals handling medical data. 
These findings underscore the enduring value of the vanilla FedAvg approach as a trusted baseline for clinical practice.

\keywords{Federated Learning  \and Resource-Constrained \and Stability.}

\end{abstract}

\section{Introduction}
\label{sec:intro}
Deep learning \cite{goodfellow2016deep,simeone2022machine} has demonstrated remarkable success across various domains by leveraging large-scale datasets for training sophisticated neural networks.
While traditional deep learning approaches assume centralized data availability, many real-world scenarios, particularly in healthcare, face privacy constraints that prevent direct data sharing \cite{abouelmehdi2018big,guan2024federated,nguyen2022federated}. 
The growing concerns about data privacy and security have led to the emergence of federated learning (FL) \cite{mcmahan2017communication}, which enables collaborative model training while keeping sensitive data localized \cite{antunes2022federated,rauniyar2023federated}. 
FL has shown particular promise in healthcare applications, where patient data privacy is paramount and regulatory compliance is strictly enforced \cite{kairouz2021advances,rieke2020future}. 
The deployment of FL in medical institutions has enabled collaborative learning across multiple hospitals without compromising patient confidentiality \cite{joshi2022federated,pfitzner2021federated}. 
However, the effectiveness of FL heavily depends on the choice of the key hyperparameters across diverse scenarios.

Recent FL techniques have developed various approaches to address client drift challenges caused by data heterogeneity across distributed clients \cite{li2019convergence,zhao2018federated}. 
FedProx \cite{li2018federated} and FedDyn \cite{acarfederated} tackle client drift through different regularization approaches - FedProx adds a proximal term to constrain local training, while FedDyn introduces dynamic correction terms to align local and global objectives. 
FedCM \cite{xu2021fedcm} takes a different approach by incorporating momentum in client training process to improve convergence stability. 
Another line of research focuses on sharpness-aware optimization \cite{foretsharpness}, starting with FedSAM \cite{qu2022generalized} which applies local perturbations to find flatter minima. 
Building upon this, FedGamma \cite{10269141} extends the concept to global optimization, FedSpeed \cite{sunfedspeed} combines it with proximal regularization for longer training intervals, and FedSMOO \cite{sun2023dynamic} integrates dynamic regularization with sharpness-awareness.
As shown in Table \ref{tab:fl_methods_opt}, these methods introduce additional optimization mechanisms either at client-side, server-side, or both, all requiring careful hyperparameter tuning.

In this paper, we demonstrate that such complexity may not be necessary, as the simple averaging mechanism of FedAvg achieves comparable or better performance.
Moreover, FedAvg remains a standard benchmark for recent FL studies, validating its enduring relevance despite newer sophisticated methods.
Indeed, the straightforward nature of FedAvg makes it less susceptible to implementation errors and easier to debug compared to more complex alternatives. 
Furthermore, we numerically show that vanilla FedAvg's simple averaging mechanism effectively captures the essential aspects of distributed learning across various scenarios. 
The consistent performance of FedAvg raises important questions about the practical value of more complex FL variants. The main contributions of this paper are as follows:

\begin{itemize}
\item We validate that FedAvg achieves comparable convergence speed to recent FL methods across communication rounds.
\item We evaluate the top-1 test accuracy of FL methods across various classification models, revealing FedAvg's consistent performance regardless of model architecture.
\item We investigate the sensitivity of FL methods to hyperparameter variations, showing that FedAvg maintains stable performance without requiring tuning.
\item We show that advanced FL methods can potentially surpass FedAvg, but finding their optimal hyperparameters is challenging.
\end{itemize}

\begin{table}[t]
\centering
\caption{Comparison of optimization techniques used in different FL methods. \xmark\ indicates the use of vanilla optimization (local SGD for $ClientOpt$ or mean aggregation for $ServerOpt$), while \cmark\ indicates the adoption of advanced optimization techniques.}
\vspace{-0.2cm}
\label{tab:fl_methods_opt}
\resizebox{\textwidth}{!}{%
        \begin{tabular}{lcccccccc}
            \toprule
            \textbf{Optimization} & \textbf{FedAvg} & \textbf{FedProx} & \textbf{FedDyn} & \textbf{FedCM} & \textbf{FedSAM} & \textbf{FedGAMMA} & \textbf{FedSpeed} & \textbf{FedSMOO} \\
            \midrule
            $ClientOpt$ & \xmark & \cmark & \cmark & \cmark & \cmark & \cmark & \cmark & \cmark \\
            $ServerOpt$ & \xmark & \xmark & \xmark & \cmark & \xmark & \cmark & \xmark & \cmark \\
            \bottomrule
        \end{tabular}%
    }
\vspace{-0.5cm}
\end{table}

The remainder of this paper is organized as follows.
Section \ref{sec:system} introduces problem setting and details the general FL process.
Section \ref{sec:experiment} validates the stability of FedAvg through comprehensive experiments.
Finally, we present our concluding remarks in section \ref{sec:conclusion}.

\section{Problem and Method}
\label{sec:system}

\subsection{Federated Setting}
We consider a federated network consisting of central server and $N$ clients. 
Each client $n = 1,\ldots,N$ holds different local dataset $\mathcal{D}^n$ with potentially varying number of data points. 
The goal of FL is to train a globally shared model $\theta \in \Theta$ by leveraging the clients' datasets $\{\mathcal{D}^n\}_{n=1}^N$ without direct data sharing. Mathematically, the training objective can be formulated as minimizing 
\begin{align}
F(\theta) \triangleq \frac{1}{N} \sum_{n=1}^{N}F_n(\theta),    
\end{align}
where $F(\theta)$ is the global objective function and $F_n(\theta)$ denotes the local objective function for the $n$th client defined as 
\begin{align}
F_n(\theta) = \frac{1}{|\mathcal{D}^n|}\sum_{x \in \mathcal{D}^n}f_n(\theta; x),
\end{align}
with $f_n(\theta; x)$ being the loss function for data sample $x$.

\begin{figure}[t]
     \centering
     \includegraphics[width=\columnwidth]{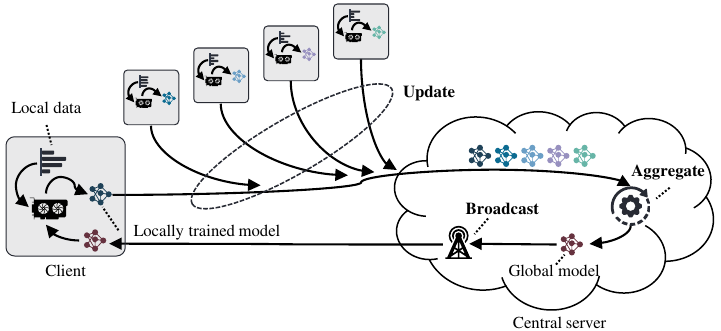}
     \caption{Overview of the general FL framework. The process consists of three main steps - (1) broadcast of global model from central server to all clients, (2) local training at randomly selected clients using their private data, and (3) aggregation of locally trained models at the central server to improve the global model.}
     \label{fig:1}
     \vspace{-0.3cm}
\end{figure}

\subsection{General FL Framework}
In FL, the central server communicates with the clients over $R$ communication rounds to minimize the objective function $F(\theta)$ as illustrated in Fig. \ref{fig:1}. The standard FL framework consists of the following steps:
\subsubsection{Broadcast}
At each communication round $r$, the central server broadcasts its aggregated model $\theta_r$ to all clients. Note that the global model is randomly initialized at the very first round ($r = 0$). 
Then the central server randomly samples a subset of $M \ll N$ clients from the total $N$ clients, denoted as $\mathcal{S}_r$, to participate in the update process.
Only the selected clients participate in the model training while the remaining clients remain idle until the next round.

\subsubsection{Update}
After receiving $\theta_r$ from the central server, each client $n$ initializes its local model as $\theta_r^n \leftarrow \theta_r$, where $\theta_{r}^n$ is the local model of the $n^{th}$ client. 
Then each randomly selected client $m \in \mathcal{S}_r$ trains its local model with $\mathcal{D}^m$ up to the maximum number of local epochs $L^m$ as follows:
\begin{align}
\theta_r^{m,L^m} \leftarrow ClientOpt(\theta_r, \mathcal{D}^m, L^m, \Phi_{client}).
\end{align}
The local training process varies depending on the FL method - vanilla FedAvg performs standard local SGD, while advanced methods incorporate additional optimization techniques with client-side hyperparameters $\Phi_{client}$.

\subsubsection{Aggregate}
After receiving the set of models $\{\theta_r^m\}_{m \in \mathcal{S}_r}$, the central server aggregates these locally trained models using ServerOpt as:
\begin{align}
\theta_{r+1} \leftarrow ServerOpt(\{\theta_r^m\}_{m \in \mathcal{S}_r}, \Phi_{server}),
\end{align}
where $ServerOpt(\cdot)$ can be any existing optimization-based FL method at the server-side.
Note that this formulation supports not only simple averaging, but also advanced optimization techniques that incorporate the server-side hyperparameters \(\Phi_{server}\) to refine the aggregated model.
The improved global model $\theta_{r+1}$ is then used for the next round $r+1$, and this process repeats until $r$ reaches the predefined number of rounds $R$ for convergence. 
Overall procedure is described in Algorithm \ref{alg:fl_framework}.

\begin{algorithm}[t]
\DontPrintSemicolon
\SetAlgoLined
\SetNlSty{}{}{:} 
\SetKwInOut{Input}{Input}\SetKwInOut{Output}{Output}

\textbf{Server Execution:}\;
\For{$r = 0,\ldots,R-1$}{
    \textbf{Broadcast:}\;
    \quad Transmit global model $\theta_r$ to all clients\;
    \quad Randomly sample $M$ clients, $\mathcal{S}_r$\;
    \textbf{Update:}\;
    \quad \textbf{For each} client $m \in \mathcal{S}_r$ \textbf{in parallel}: \\
    \quad \quad $\theta_r^{m,L^m} \leftarrow ClientOpt(\theta_r, \mathcal{D}^m, L^m, \Phi_{client})$\;
    \textbf{Aggregate:}\;
    \quad $\theta_{r+1} \leftarrow ServerOpt\Big(\{\theta_r^m\}_{m \in \mathcal{S}_r}, \Phi_{server}\Big)$\;
}
\caption{General FL Framework.}
\label{alg:fl_framework}
\end{algorithm}

\section{Experiment and Results}
\label{sec:experiment}

\subsection{Experiment Setting}
To validate the stability of FedAvg, we evaluate the performance of state-of-the-art FL methods on two medical image classification tasks: blood cell \cite{acevedo2020dataset} and skin lesion \cite{tschandl2018ham10000} classification task. 
For all experiments except the model comparison experiment, clients employ Vision Transformer (ViT) \cite{dosovitskiy2020image} as local model to ensure consistent evaluation of the FL methods.
Moreover, we adopt label skew \cite{li2022federated} to introduce data heterogeneity by distributing the non-IID data over $N=100$ clients following a Dirichlet distribution ($\alpha=0$).
Additionally, $M=10$ clients are randomly selected at each communication round to participate in the training process.
All experiments are conducted on an Intel Gaudi 2 AI accelerator to facilitate efficient, high-performance learning. 

\subsection{Results}
\begin{figure}[t]
\centering
\includegraphics[width=\columnwidth]{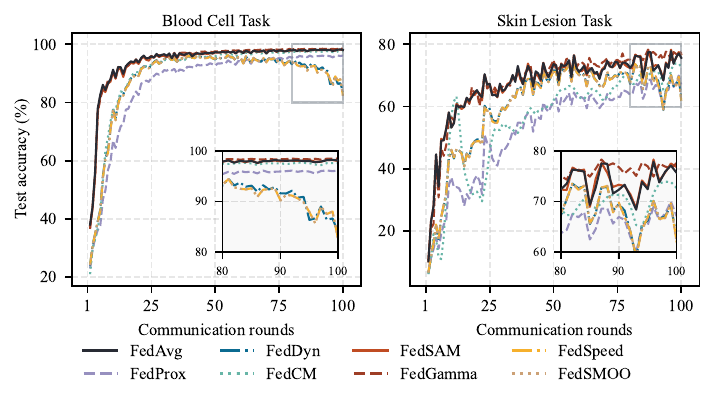}
\vspace{-1cm}
\caption{Test accuracy vs. communication rounds for blood cell and skin lesion classification tasks. FedAvg shows comparable convergence speed and performance to state-of-the-art FL methods. Zoom-in plots highlight the comparable and stable performance of vanilla FL during final rounds.}
\label{fig:result1}
\vspace{-0.5cm}
\end{figure}

\subsubsection{Impact of Stability}
To evaluate the stability of different FL methods, we analyze the test accuracy with respect to communication rounds on both blood cell and skin lesion classification tasks as shown in Fig. \ref{fig:result1}. 
The results demonstrate that FedAvg achieves comparable or even faster convergence compared to more sophisticated FL methods such as FedProx, FedCM, FedDyn, and FedSpeed. 
Specifically, FedAvg exhibits stable convergence patterns, reaching competitive test accuracies within 100 communication rounds for both datasets. 
The results challenge the common belief that advanced FL methods necessarily provide faster convergence.

Furthermore, the zoom-in plots during the final 20 rounds reveal that FedAvg maintains competitive performance in the convergence phase, suggesting that the complexity of recent FL methods may not translate to meaningful improvements in convergence speed without tuning. 
The consistent performance of FedAvg across both medical imaging tasks indicates its robust convergence properties despite its simplicity. 
This empirical evidence demonstrates that the straightforward averaging mechanism of FedAvg effectively captures the essential aspects of collaborative learning, making it a reliable choice for medical AI tasks.

\begin{figure}[t]
\centering
\includegraphics[width=\columnwidth]{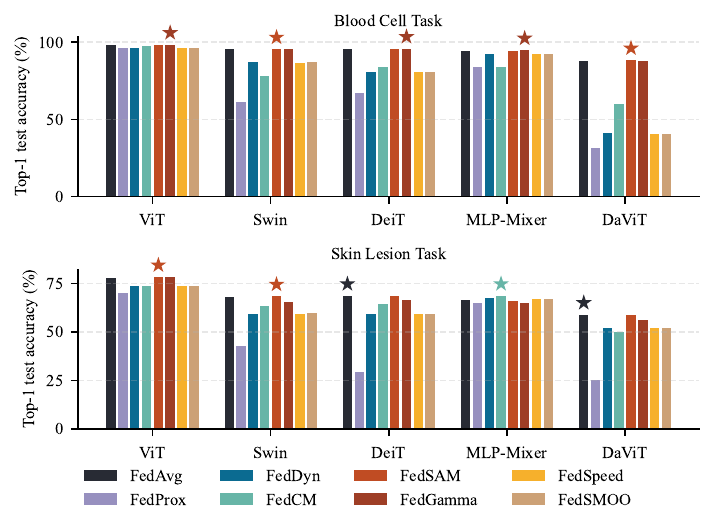}
\vspace{-1cm}
\caption{Comparison of top-1 test accuracy across different model architectures. Results show that FedAvg maintains stable performance regardless of the underlying model on both blood cell and skin lesion classification tasks. $\bigstar$ denotes the best performance.}
\label{fig:result2}
\vspace{-0.5cm}
\end{figure}
\vspace{-0.5cm}
\subsubsection{Impact of Model Diversity}
To investigate how different model architectures affect FL performance, we evaluate the top-1 test accuracy across representative vision models: ViT, DeiT \cite{touvron2021training}, DaViT \cite{ding2022davit}, Swin \cite{liu2021swin}, and MLP-Mixer \cite{tolstikhin2021mlp}. 
As shown in Fig. \ref{fig:result2}, for the blood cell classification task, FedGamma achieved the best performance on ViT, DeiT, and MLP-Mixer, while FedSAM led on Swin and DaViT.
In the skin lesion classification task, FedAvg delivered the top results on DeiT and DaViT, FedSAM excelled on ViT and Swin, and FedCM attained the maximum performance on MLP-Mixer. 
Notably, even when FedAvg was not the best, its performance was nearly indistinguishable from the top method.
Therefore, in resource-constrained environments, FedAvg is the cost-effective choice for achieving reasonable results without the need for hyperparameter tuning.

\begin{table}[t]
\centering
\footnotesize
\setlength{\tabcolsep}{3pt}
\caption{Performance comparison of different FL methods on blood cell and skin lesion classification tasks. Results show that advanced FL methods exhibit significant performance variations across different hyperparameter settings, while FedAvg (marked with \colorbox{gray!20}{gray}) maintains consistent performance without any tuning. The time costs for each FL method are measured when executed independently on a single Gaudi 2 chip.}
\label{tab:comparison}
\begin{tabularx}{\textwidth}{@{}l l c c c c c c@{}}
\toprule
Method & Key $\Phi$ & \multicolumn{3}{c}{Blood Cell Task} & \multicolumn{3}{c}{Skin Lesion Task} \\
\cmidrule(lr){3-5}\cmidrule(lr){6-8}
& & IID & non-IID & Time/Round & IID & non-IID & Time/Round \\
\midrule
\rowcolor{gray!20} 
FedAvg
& \multicolumn{1}{c}{-} 
& 98.45 & 98.25 & 31.04s
& 81.60 & 77.86 & 24.98s \\
\midrule
\multirow{3}{*}{FedProx}
& $\lambda=0.1$
& 96.52 & 96.14 & \multirow{3}{*}{31.91s}
& 74.86 & 69.98 & \multirow{3}{*}{25.39s} \\
& $\lambda=0.01$
& 98.39 & 98.13 &
& 81.40 & 77.96 & \\
& $\lambda=0.001$
& 98.42 & 98.25 &
& 81.65 & 77.91 & \\
\midrule
\multirow{3}{*}{FedDyn}
& $\beta=0.1$
& 96.58 & 96.23 & \multirow{3}{*}{33.28s}
& 77.21 & 73.52 & \multirow{3}{*}{26.96s} \\
& $\beta=0.01$
& 96.58 & 96.23 &
& 77.21 & 73.52 & \\
& $\beta=0.001$
& 96.58 & 96.23 &
& 77.21 & 73.52 & \\
\midrule
\multirow{3}{*}{FedCM}
& $\mu=0.1$
& 97.84 & 97.72 & \multirow{3}{*}{32.54s}
& 75.41 & 73.92 & \multirow{3}{*}{26.38s} \\
& $\mu=0.01$
& 87.99 & 87.46 &
& 66.93 & 50.07 & \\
& $\mu=0.001$
& 32.42 & 30.02 &
& 66.93 & 25.69 & \\
\midrule
\multirow{3}{*}{FedSAM}
& $\rho=0.1$
& 93.13 & 85.03 & \multirow{3}{*}{45.40s}
& 70.17 & 68.43 & \multirow{3}{*}{32.11s} \\
& $\rho=0.01$
& 98.48 & 98.13 &
& 82.00 & 78.05 & \\
& $\rho=0.001$
& 98.42 & 98.22 &
& 81.70 & \textsb{\underline{78.30}} & \\
\midrule
\multirow{3}{*}{FedGamma}
& $\rho=0.1$
& 92.17 & 80.06 & \multirow{3}{*}{45.75s}
& 67.83 & 67.23 & \multirow{3}{*}{33.87s} \\
& $\rho=0.01$
& 98.60 & \textsb{\underline{98.54}} &
& 82.89 & 76.61 & \\
& $\rho=0.001$
& \textsb{\underline{98.68}} & 98.51 &
& \textsb{\underline{83.74}} & \textsb{\underline{78.30}} & \\
\midrule
\multirow{3}{*}{FedSpeed}
& $\rho=0.1$
& 37.27 & 42.09 & \multirow{3}{*}{45.13s}
& 66.88 & 67.43 & \multirow{3}{*}{34.28s} \\
& $\rho=0.01$
& 96.49 & 95.97 &
& 75.06 & 72.82 & \\
& $\rho=0.001$
& 96.64 & 96.26 &
& 76.96 & 73.67 & \\
\midrule
\multirow{3}{*}{FedSMOO}
& $\rho=0.1$
& 35.78 & 40.13 & \multirow{3}{*}{53.68s}
& 66.88 & 67.53 & \multirow{3}{*}{40.68s} \\
& $\rho=0.01$
& 96.49 & 95.97 &
& 75.11 & 72.77 & \\
& $\rho=0.001$
& 96.64 & 96.26 &
& 77.06 & 73.67 & \\
\bottomrule
\end{tabularx}
\vspace{-1cm}
\end{table}

\subsubsection{Impact of Hyperparameter Selection}
We examine hyperparameter sensitivity under both IID and non-IID data by comparing the performance of vanilla FL and state-of-the-art methods. 
As presented in Table \ref{tab:comparison}, FedProx is highly sensitive to the $\lambda$, showing substantially lower performance at $\lambda=0.1$ than at $\lambda=0.001.$ Unlike FedProx, FedDyn demonstrates nearly uniform accuracy across different $\beta$ values, implying minimal impact of its correction term within a certain range. 
Meanwhile, FedCM converges unstably at $\mu=0.001,$ recording only about $30\%$ accuracy on the blood cell classification task. 
Nextly, FedSAM and FedGamma undergo considerable performance drops at $\rho=0.1,$ underscoring their sensitivity to aggressive sharpness exploration. 
Similarly, FedSpeed and FedSMOO exhibit reduced initial accuracy, stabilizing only at smaller $\rho$ values. These methods also demand more time per round than FedAvg, incurring larger communication overhead.

In contrast, FedAvg consistently surpasses 98\% (blood cell) and 77\% (skin lesion) without additional refinement across both IID and non-IID settings. 
Its straightforward design eliminates the need for tuning complicated hyperparameters, thus saving both time and computational resources.
Consequently, the reliability and efficiency of FedAvg make it a more practical choice for real-world medical AI applications in resource constrained settings, where achieving maximum impact at minimal cost is essential.

\begin{table}[t]
  \centering
  \caption{Comparison of convergence speed and test accuracy across FL methods using their sub optimal hyperparameters. Results show that even with optimal settings, advanced FL methods achieve marginal improvements over FedAvg (marked with \colorbox{gray!20}{gray}) at the cost of hyperparameter tuning.}
  \label{tab:ablation}
  \begin{tabularx}{\columnwidth}{l *{4}{>{\centering\arraybackslash}X}}
    \toprule
    \multirow{2}{*}{Method} & \multicolumn{2}{c}{Blood Cell Task} & \multicolumn{2}{c}{Skin Lesion Task} \\
    \cmidrule(lr){2-3} \cmidrule(lr){4-5}
     & \mbox{Top-1 Test Acc. (\%)} & Round & \mbox{Top-1 Test Acc. (\%)} & Round \\
    \midrule
    \rowcolor{gray!20} 
    FedAvg    & 98.25 & 99  & 77.86  & 96  \\
    FedProx   & 98.25 & 99  & 77.96  & 96  \\
    FedDyn    & 96.23 & 55  & 73.52  & 82  \\
    FedCM     & 97.72 & 93  & 73.92  & 98  \\
    FedSAM    & 98.22 & 98  & 78.30  & 96  \\
    FedGamma  & 98.54 & 98  & 78.30  & 87  \\
    FedSpeed  & 96.26 & 55  & 73.67  & 56  \\
    FedSMOO   & 96.26 & 55  & 73.67  & 56  \\
    \bottomrule
  \end{tabularx}
  \vspace{-0.5cm}
\end{table}

\subsubsection{Ablation Study}
To evaluate state-of-the-art FL methods under their sub optimal settings, we analyze their convergence behavior using the best-performing hyperparameters from our previous experiments. 
As shown in Table \ref{tab:ablation}, while some methods like FedGamma achieve marginally better accuracy and faster convergence compared to FedAvg, others show similar or degraded performance despite optimal settings. 
Methods like FedDyn, FedSpeed, and FedSMOO converge faster but with lower accuracy, while FedProx and FedSAM show comparable performance to FedAvg. 
These results indicate that advanced FL methods may outperform FedAvg, but finding optimal hyperparameters remains a key challenge requiring further investigation.

\section{Conclusion}
\label{sec:conclusion}
In this paper, we conducted a comprehensive study on the stability of FedAvg against advanced FL methods across medical image classification tasks. 
Our numerical results show that vanilla FL maintains robust performance under diverse conditions without requiring hyperparameter tuning. 
Moreover, we have shown that some advanced FL methods can marginally outperform FedAvg with their optimal settings, but the extensive effort required for hyperparameter tuning may not justify these modest gains. 
These findings suggest that FedAvg's algorithmic simplicity is its strength, making it a reliable and efficient choice for practical medical FL deployments where resources are constrained.

%
%
%
\bibliographystyle{splncs04}
\bibliography{reference}

\end{document}